# Scorpion detection and classification systems based on computer vision and deep learning for health security purposes

**Francisco Luis Giambelluca[1], Marcelo A. Cappelletti[1,2,3], Jorge Osio[1,2], Luis A. Giambelluca[3,4]**

[1] Grupo de Control Aplicado (GCA), Área CeTAD, Instituto LEICI (CONICET-UNLP). Calle 116 y 48, 2° piso, La Plata (1900), Argentina.

[2] Programa TICAPPS, Universidad Nacional Arturo Jauretche, Av. Calchaquí 6200, Florencio Varela (1888), Argentina

[3] Consejo Nacional de Investigaciones Científicas y Técnicas (CONICET), Argentina

[4] CEPAVE (CONICET-UNLP-CCT La Plata, Asoc. CICBA). Boulevard 120 e/61 y 64, La Plata (1900), Argentina.

E-mail: francisco.giambelluca@ing.unlp.edu.ar; marcelo.cappelletti@ing.unlp.edu.ar; jorge.osio@ing.unlp.edu.ar; giambelluca@cepave.edu.ar

**Abstract**

In this paper, two novel automatic and real-time systems for the detection and classification of two genera of scorpions found in La Plata city (Argentina) were developed using computer vision and deep learning techniques. The object detection technique was implemented with two different methods, YOLO (You Only Look Once) and MobileNet, based on the shape features of the scorpions. High accuracy values of 88% and 91%, and high recall values of 90% and 97%, have been achieved for both models, respectively, which guarantees that they can successfully detect scorpions. In addition, the MobileNet method has been shown to have excellent performance to detect scorpions within an uncontrolled environment and to perform multiple detections. The MobileNet model was also used for image classification in order to successfully distinguish between dangerous scorpion (*Tityus*) and non-dangerous scorpion (*Bothriurus*) with the purpose of providing a health security tool. Applications for smartphones were developed, with the advantage of the portability of the systems, which can be used as a help tool for emergency services, or for biological research purposes. The developed systems can be easily scalable to other genera and species of scorpions to extend the region where these applications can be used.

## 1. Introduction

For many years scorpions have been object of study, due to their high danger [1]. For this reason, the need for their detection and identification is raised. Usually, the scorpion bites happen accidentally, which is due to the ignorance of the presence of these arachnids.

Scorpions are nocturnal animals with negative phototropism. They can be found sheltered during the day, in natural environments, under rocks or inside holes, although some species can also be domiciliary, which remain hidden in the daytime and roam at night.

Argentina has only one genus (*Tityus*) considered of sanitary importance, since it causes accidents every year, and some can be fatal, especially in children and elderly. This genus has several highly dangerous species, such as *Tityus confluence*, *Tityus serrulatus*, *Tityus bahiensis* and *Tityus trivittatus*. Other species of this genus can cause great pain with their bites, although they do not register fatal accidents. Specifically, in the La Plata district, there are two species of scorpions: *Tityus trivittatus* and *Bothriurus bonaeriensis* (Fig. 1). The first one is of sanitary importance and is mainly located in urban areas, whereas the second one is not dangerous for humans and is commonly found in rural and peri-urban areas, such as can be seen in Fig. 2.

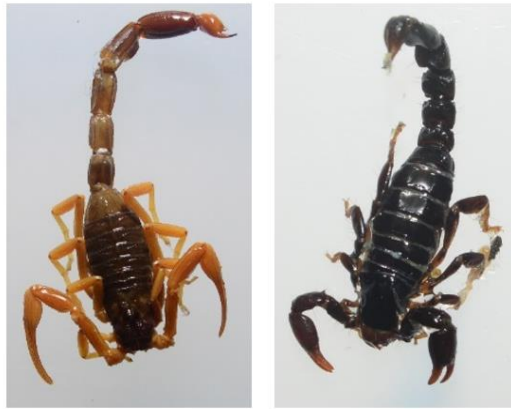

**Fig. 1.** Pictures of Tityus trivittatus (left) and Bothriurus bonaeriensis (right).

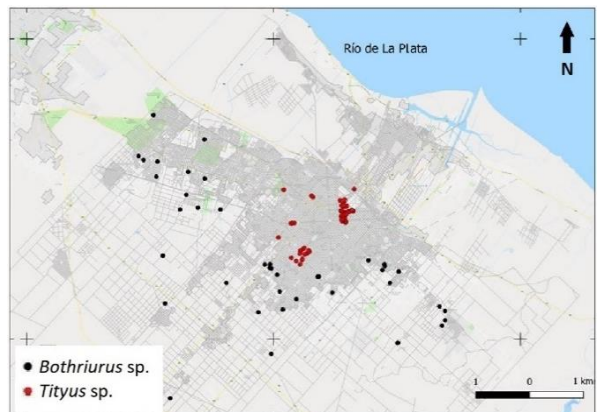

**Fig. 2.** Distribution of scorpions in La Plata district (Argentina).

In recent years, the number of consultations carried out at the CEPAVE Arachnology Laboratory (CONICET-UNLP- Assoc. CICPBA) have increased substantially, which is due to the increase in the appearance of scorpions in different areas of the La Plata city (Argentina). The query species are mainly *Bothriurus bonariensis* and *Tityus trivittatus* [2].

Although *Tityus trivittatus* and *Bothriurus bonaeriensis* have many similarities, they also have some differences, especially concerning their morphology, in the shape of their tails and pedipalp. The difference in dangerousness between these species makes it necessary to identify them correctly, in order to avoid determination errors in case of accidents, and thus proceed according to the dangerousness of the scorpion.

The oldest methods of scorpion detection include rock rolling, burrowing detection, peeling loose back of tree and pitfall trap, which are dangerous, time consuming and invasive [3,4]. Other methods of detection use different biological characteristics of scorpions such as vibration signals or the fluorescent property of scorpions [5,6].

Machine learning (ML) is a branch of artificial intelligence whose objective is to develop techniques that allow computing units to learn [7]. These techniques are based on algorithms, which convert databases into well-defined classifiers, without having to supervise the development of the latter. On the other hand, computer vision is a set of computer techniques developed in order to interpret and process digital images, and to imitate (and even improve) the human vision system [8]. In recent years, aiming to improve their performance, ML techniques have begun to be used in the field of computer vision, for example, to detect, recognize and classify objects [9–12], to face recognition [13], to analyze textures [14], to arm fracture detection [15], image recovery [16], among other applications.

The authors have previously developed an automatic system for the detection and classification of scorpions using computer vision and ML approaches [17]. On the one hand, the fluorescent property of scorpions when exposed to ultraviolet light and their shape features were used as a double validation process for the detection system of these arachnids through image processing techniques; and, on the other hand, three different models based on ML algorithms used to discriminate between two genera of scorpions (*Bothriurus* and *Tityus*) were compared. Specifically, they are the local binary-pattern histogram (LBPH) algorithm and deep neural networks with transfer learning (DNN with TL) and data augmentation (DNN with TL and DA) approaches.

In order to improve the results obtained with these models, in this work we present the development of two systems of automatic detection and classification of scorpions in real time using different deep learning models, to those used in [17]. On the one hand, object detection was implemented with two different models, based on the shape features of the scorpions, which were compared with each other. Both models, known as YOLO (You Only Look Once [18]) and MobileNet [19] with TensorFlow platform, are based on deep convolutional neural networks [20] for real-time object detection. On the other hand, the MobileNet model with TensorFlow was used for image classification. In particular, we have used the fourth version of YOLO (*YOLOv4*) for object detection [21], and the second version of MobileNet (*MobileNetV2*) [22] for object detection and image classification systems. Fig. 3 shows the flow chart of the systems under study in this paper.

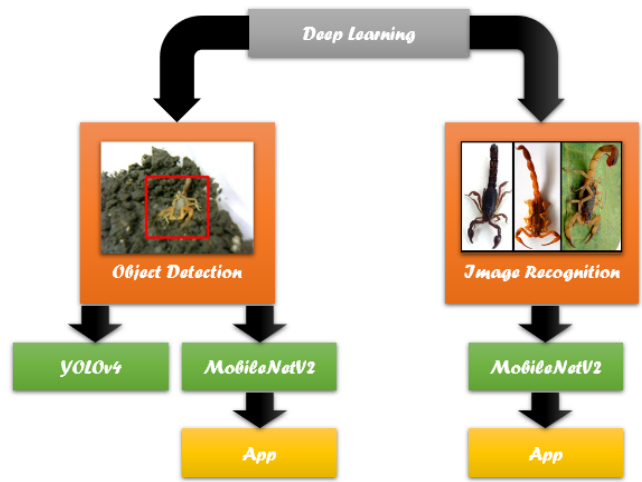

**Fig. 3.** Schematic flow chart of the systems studied in this work.

The data set used was extracted from the database belonging to the CEPAVE Arachnology Laboratory. A selection of the best images was made. Also, new images acquired by photographing all the specimens of the scorpion collection of the previously mentioned Laboratory were added to the data set.

In the object detection systems, a total of 612 images of scorpions (positive images) of very good quality and sharpness were considered. To reduce the false positive rate, that is, detection in the absence of the target, 197 images without scorpions (negative images) were added. Additionally, Roboflow [23] was used to increase the database, reaching a total of 1937 images, which were randomly distributed to training, validation, and test datasets in a 70:20:10 ratio, respectively. Since the training process involves the training and validation stages, there is a 77:23 relationship, which is very close to the "Pareto principle" of 80:20 rule [24].

The database for the training purpose was exported using the TFRecord format, and a link was generated to be able to use it directly from Google Colaboratory. The image editing values for the database augmentation were as follows:

- Flip (horizontal and vertical)

- 90° Rotation (both ways)
- 45° Rotation (both ways)
- 45° Tilt (both wayss)
- 48% Saturation (plus and minus)
- 25% Exposure (plus and minus)
- Blur (1.75pX)
- Noise (5%)

In order to validate and better understand the scope of the trained models, confusion matrix were implemented to classify the results and evaluate the performance of these models from the following four metrics: accuracy (A), precision (P), recall (R), and $F_{measure}$ (F1). The following equations are used to calculate these metrics [25]:

$$A = \frac{(TP+TN)}{(TP+TN+FP+FN)} \quad (1)$$

$$P = \frac{TP}{(TP+FP)} \quad (2)$$

$$R = \frac{TP}{(TP+FN)} \quad (3)$$

$$F_{measure} = 2 \cdot \frac{P.R}{(P+R)} \quad (4)$$

where TP, TN, FP and FN denote true positives, true negatives, false positives and false negatives, respectively.

In both object detection systems, the presence of the scorpion in the image was considered as a true positive case (TP). Then, the worst situation occurs when the system fails to detect the scorpion, due to the health risk that this entails. That is, in the case of absence of detection when the effective presence of a scorpion was omitted (case of false negative (FN)). Therefore, since R increases when FN decreases, recall is the metric of greatest interest in the developed detection systems.

The receiver operating characteristic (ROC) curve [26] was also used to examine the behavior of the binary model when the detection threshold is changed. Only the TPs and FPs are necessary to graph the ROC curve, in which the TP rate (sensitivity) is plotted against the FP rate (1-specificity). Every point in a ROC space corresponds to a given instance of the confusion matrix and represents a relative trade-off between TP and FP. The higher the values of TP with respect to FP, the better the trained model will be. The optimal model corresponds to the point located in the upper left corner of the ROC space, with coordinates (0,1). which represents 100% sensitivity (no false negatives) and 100% specificity (no false positives).

On the other hand, the image classification system was implemented using the MobileNetV2 model from the web environment "teachable machine" [27] provided by Google. The database was divided by genus in three categories: Tityus with 105 images, Bothriurus with 113 images, and "None" (for the absence of a scorpion) with 60 images. Roboflow was used to increase the database, reaching a total of 315, 339, and 180 images, respectively, for each category.

For both object detection and image classification systems, applications for smartphones were developed. Today, these devices have the necessary computing capacity for such activities, with the advantage of the portability of the systems. These applications can be used for help tool for emergency services, biological research, and civil use, among others. Both smartphone applications developed in this paper were implemented in the TensorFlow Lite model, which was obtained by making the corresponding transformations to the trained models. Also, the developed systems in this work are easily scalable to other genera and species of scorpions to extend the region where these applications can be used.

## 2. Implementation

*2.1. Object Detection*

*2.1.1. YOLOv4*

In this detection system, the "Scaled-YOLOv4" [28] was used for the training and testing phases, which provides the necessary tools. The images used for the training process have a size of 416x416 pixels, the batch size was set to 16, and the network was trained for 600 epochs. The calculations were run in the free access web development environment "Google Colaboratory". A Tesla T4 GPU with 15GB allocated memory was used. The processing time for training was close to 8 hours. Compared to other similar systems, this system is increasingly used for its high speed and efficiency.

In Fig. 4, the progress of the precision and recall metrics can be observed during training. Both metrics are saturated at a value close to 70%, which ensures a complete training with the database used, and without overtraining. It can also be seen that recall is always above precision, which is ideal for our system that demands a low rate of FN.

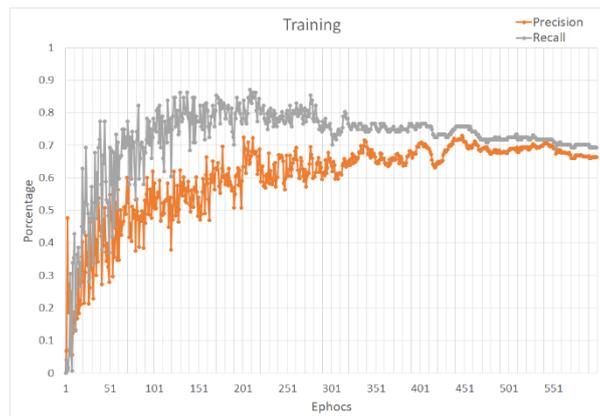

**Fig. 4.** Precision and Recall while training epochs with the YOLOv4 model.

*2.1.2. MobileNetV2*

The TensorFlow Lite model was used in this case, in order to run the detection system on a smartphone. Also, the transfer learning technique [29] was adopted for the training phase, in which the MobileNetV2 pre-trained model was used with a batch size of 12, 100 steps per epoch and a total of 400000 epochs. The processing time for training was close to four days. Fig. 5 shows the training loss as a function of the number of epochs. The performance of the system is presented in the Results section.

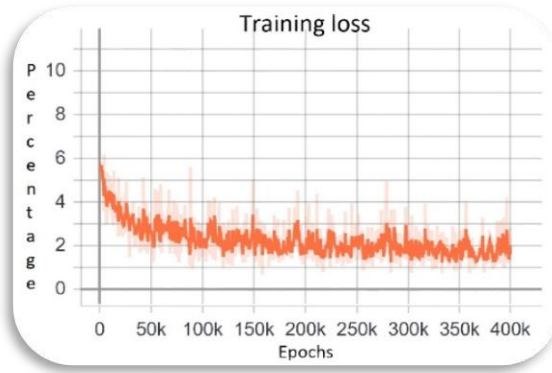

**Fig. 5.** Training loss of MobileNetV2.

*2.2. Image Classification*

The image classification system was developed with the purpose of providing a health security tool. This application consists of a classifier that discriminates between three classes: dangerous scorpion (*Tityus*), non-dangerous scorpion (*Bothriurus*), and the absence of both (an environment without the close presence of scorpions).

Teachable Machine was used for the training stage, which is based on the MobileNetV2 pre-trained model. The training consisted of 200 epochs (Fig. 6), with a batch size of 512, and a learning rate of 0.001. During this training, the expected improvement in accuracy was achieved, with a strong correlation between the training and its validation. It can also be seen that both curves are saturated at a value higher than 90%, so our system is not undertrained or overtrained.

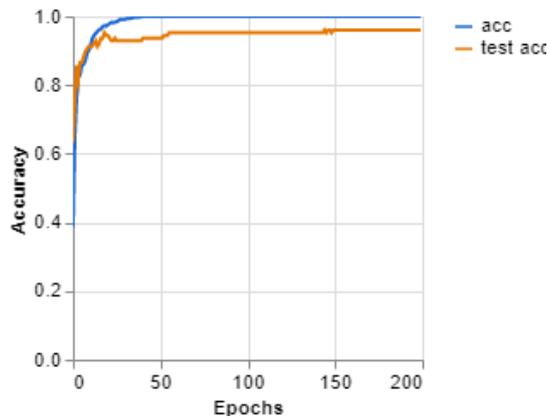

**Fig. 6.** Training and test accuracy of Image classification.

Regarding the loss function, Fig. 7 shows the expected decrease during training, reaching values close to 0% and 20% for the training and validation phases, respectively.

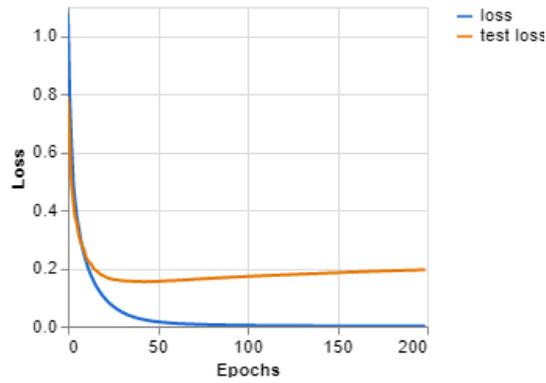

**Fig. 7.** Training and test loss of image classification.

From figures 6 and 7, it is clear that the training was achieved with acceptable accuracy and loss in order to obtain an appropriate image classification. In the Results section, the test results of this system are analysed.

## 3. Results and discussion

*3.1. Object Detection*

In this section, the performance of both object detection systems developed in this work, using the YOLOv4 and MobileNetV2 models, are compared. The behaviors of these detection systems were evaluated from 81 images of the database, which were not used in the previous training processes. These images were randomly selected through the Roboflow.

Fig. 8 shows proper scorpion detections using the YOLOv4 (left picture) and the MobileNetV2 (right picture) models. Also, a high accuracy in the detection of the scorpion (98.96%) can be seen for the MobileNetV2 model.

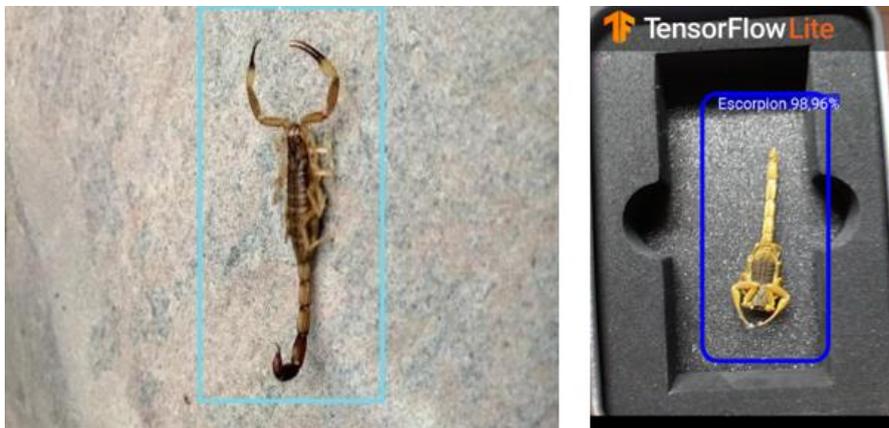

**Fig. 8.** Scorpion detection with YOLOv4 (left) and with MobileNetV2 (right) models.

The confusion matrices obtained during the testing for the YOLOv4 and MobileNetV2 models are shown in figures 9 and 10, respectively, where the vertical axes correspond to the true data and the horizontal axes correspond to the predictions of the models.

Table 1 shows the values of accuracy, precision, recall and $F_{measure}$ calculated using equations (1)–(4), respectively, for both models considered in this study. These results show that both object detection systems are able to successfully detect scorpions. Furthermore, the high values of recall obtained indicate that there are very low values of false negatives, which is essential for health security purposes. In particular, the recall of the MobileNetV2 model (0.97) is greater than that obtained by the YOLOv4 model (0.90), which implies a safer health system.

**Table 1**
Metrics for the two detection models under study

| Method | Accuracy (A) | Precision (P) | Recall (R) | $F_{measure}$ |
|---|---|---|---|---|
| YOLOv4 | 0.88 | 0.93 | 0.90 | 0.92 |
| MobileNetV2 | 0.91 | 0.92 | 0.97 | 0.94 |

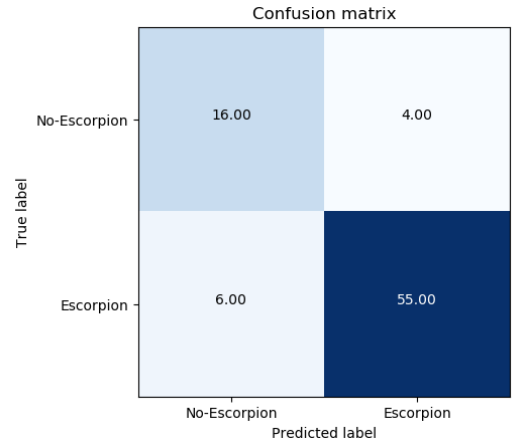

**Fig. 9.** YOLOv4 Confusion matrix.

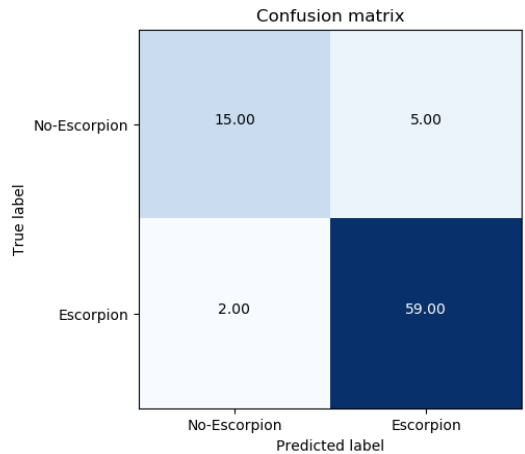

**Fig. 10.** MobileNetV2 Confusion matrix.

Fig. 11 shows the ROC curves for these detection systems. It can be seen in this figure that the areas under the ROC curves for the YOLOv4 and MobileNetV2 models are very similar, with area of 85% and 86%, respectively, which implies a very good sensitivity and specificity relationship.

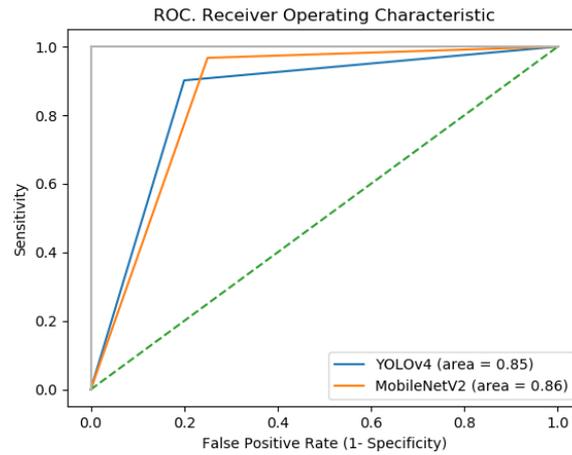

**Fig. 11.** Comparative ROC curves for the two detection models under study.

From the results obtained, it has been found that the MobileNetV2 method has better performance than YOLOv4 method for the detection of scorpions. Although the MobileNetV2 method requires a greater number of epochs and a longer training time, the comparison between both methods is valid, because both systems were trained until their saturation without overtraining.

The MobileNetV2 method has also demonstrated an excellent ability to correctly detect scorpions within an uncontrolled environment, that is, with multiple positions not present in the original database, and with various objects that make detection difficult, as shown in figure 12. It can be seen in this figure that despite the presence of the wires, and of darker and brighter areas, this system detects the scorpion correctly, and does not erroneously detect another object.

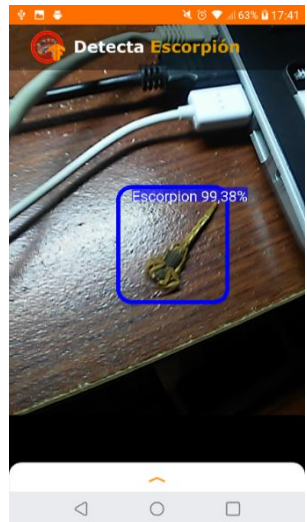

**Fig. 12.** Scorpion detection using MobileNetV2 method within an uncontrolled environment.

Additionally, the MobileNetV2 method also has an excellent responsiveness to carry out multiple detections continuously without problems, as shown in figure 13. In this image, it can be observed the presence of three scorpions of two different genera, which were all properly detected as expected.

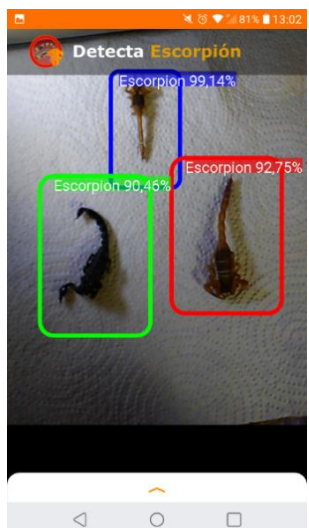

**Fig. 13.** Detection of three scorpions using MobileNetV2 method.

Finally, this system was used to detect pseudoscorpions, also known as false scorpions, which are arachnids with the general appearance of scorpions except that they have no tails. Figure 14 shows that the MobileNetV2 method mistakenly detects the pseudoscorpion as a scorpion, with 90% certainty. However, this error is more than acceptable and understandable because the trained model has never seen pseudoscorpions before.

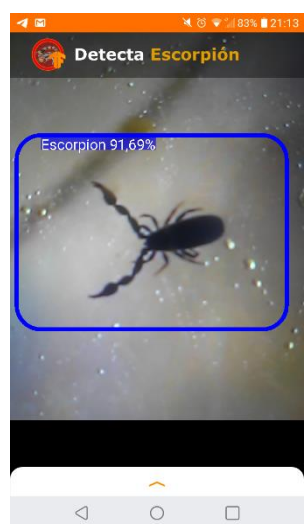

**Fig. 14.** Pseudo-scorpion detection as real scorpion using MobileNetV2 method.

*3.2. Image Classification*

In this section, the performance of the image classification system developed in this paper, using the MobileNetV2 model, is analyzed. The behavior of this system was evaluated from 126 images (15% of the database), which were not used in the previous training processes.

As has been previously mentioned, this image classification system is a classifier that discriminates between three classes: dangerous scorpion (*Tityus*), non-dangerous scorpion (*Bothriurus*), and the absence of both. Figure 15 shows the confusion matrix of 3x3 obtained during the testing, where it is possible to observe the low rate of error that is obtained in the classification process. Table 2 shows the values of accuracy, precision, recall and $F_{measure}$. These results show that our model is able to successfully discriminate between the three classes.

Since the main interest of the developed classification system is to provide a health security tool, it is essential to guarantee the correct classification of the genus *Tityus* due to its dangerousness. This condition is satisfactorily fulfilled by the proposed system.

**Table 2**
Metrics for the three classes considered in the image classification system

| Class | Accuracy (A) | Precision (P) | Recall (R) | $F_{measure}$ |
|---|---|---|---|---|
| Tityus | 0.97 | 0.96 | 0.96 | 0.96 |
| Bothriurus | 0.96 | 0.96 | 0.94 | 0.95 |
| None | 0.99 | 1.00 | 0.96 | 0.98 |

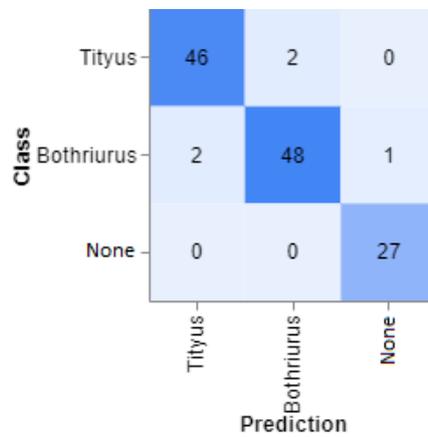

**Fig. 15.** Image Classification Confusion matrix.

The ROC curve is shown in figure 16, which indicates the very good performance of this classification system.

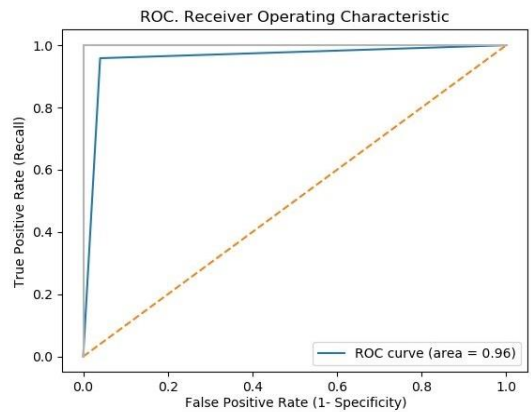

**Fig. 16.** Image Classification ROC curve.

Considering only the images corresponding to the two genera of scorpions (*Bothriurus* and *Tityus*), it is possible to compare the results obtained in this work with those presented in our previous work for three different ML approaches [17]. Table 3 summarizes the results of the four metrics considered in this paper. These results show that the MobileNetV2 method is clearly more efficient than the others to discriminate between dangerous scorpion (*Tityus*) and non-dangerous scorpion (*Bothriurus*).

**Table 3**
Comparison of metrics between the MobileNetV2 model and those presented in [17]

| Method | Accuracy (A) | Precision (P) | Recall (R) | $F_{measure}$ |
|---|---|---|---|---|
| LBPH | 0.65 | 0.78 | 0.63 | 0.70 |
| DNN with TL | 0.78 | 0.77 | 0.93 | 0.84 |
| DNN with TL and DA | 0.78 | 0.79 | 0.89 | 0.84 |
| MobileNetV2 | 0.96 | 0.96 | 0.96 | 0.96 |

Figure 17 shows the correct recognition and classification of non-dangerous (*Bothriurus*) and dangerous (*Tityus*) genera of scorpions, using the MobileNetV2 model. For example, in the right picture, this application reports that the displayed object is classified as a dangerous scorpion ("peligroso" in Spanish). Therefore, a functional system was developed to provide health security when encountering a scorpion.

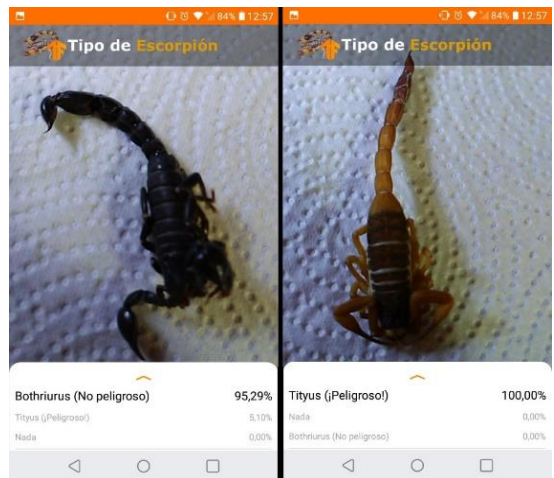

**Fig. 17.** Example of the correct classification of the two genera of scorpions: Bothriurus (left image) and Tityus (right image).

## 4. Conclusion

Novel automatic and real-time detection and classification systems capable of detecting and identifying scorpions found in La Plata city (Argentina) were proposed in this work, using computer vision and deep learning techniques. YOLOv4 and MobileNetV2 models were used and compared for object detection process, which provide high accuracy values of 88% and 91%, and high recall values of 90% and 97%, respectively. Although both object detection methods can successfully detect scorpions, the performance of the MobileNetV2 method was better than that of the YOLOv4 method. Additionally, the MobileNetV2 method also has an excellent responsiveness to detect scorpions within an uncontrolled environment and to carry out multiple detections. On the other hand, the MobileNetV2 method was also used for image classification with the purpose of providing a health security tool. The results obtained have allowed us to successfully identify between three classes: dangerous scorpion (*Tityus*), non-dangerous scorpion (*Bothriurus*), and the absence of both. A comparison between this method with those presented in our previous work has been performed, which allows us to conclude that the MobileNetV2 method is the most efficient of all the compared methods. Furthermore, for both object detection and classification systems, applications for smartphones were developed. These applications capable of detecting and classifying scorpions allow the portability of the systems and protect people from dangerous animals.

**Acknowledgements**

This work was partially supported by the Universidad Nacional de La Plata, Argentina and by the Research National Council (CONICET), Argentina.